\documentclass{article}

\usepackage{arxiv}

\usepackage[utf8]{inputenc} 
\usepackage[T1]{fontenc}    
\usepackage{hyperref}       
\usepackage{url}            
\usepackage{booktabs}       
\usepackage{amsfonts}       
\usepackage{nicefrac}       
\usepackage{microtype}      
\usepackage{graphicx}
\usepackage{natbib}
\usepackage{doi}
\usepackage{amssymb}
\usepackage{amsmath}
\usepackage{amsthm}
\usepackage{mathrsfs}
\usepackage{float} 
\usepackage[normalem]{ulem}
\usepackage{tabularray}
\usepackage{algorithm}
\usepackage{caption}
\usepackage{algpseudocode}
 \newtheorem{remark}{Remark}

\title{DGNN: A Neural PDE Solver Induced by Discontinuous Galerkin Methods}

\author{ \hspace{1mm}Guanyu Chen \\
	College of Integrated Circuits\\
	Zhejiang Uinversity\\
	\texttt{guanyu@zju.edu.cn} \\
	\And
	\hspace{1mm}Shengze Xu \\
	Department of Mathematics\\
	The Chinese University of Hong Kong\\
	\texttt{szxu@math.cuhk.edu.hk} \\
        \And
	\hspace{1mm}Dong Ni\thanks{Corresponding author} \\
	College of Integrated Circuits\\
	Zhejiang Uinversity\\
	\texttt{dni@zju.edu.cn} \\
        \And
	\hspace{1mm}Tieyong Zeng\\
	Department of Mathematics\\
	The Chinese University of Hong Kong\\
	\texttt{zeng@math.cuhk.edu.hk} \\
}

\hypersetup{
pdftitle={A template for the arxiv style},
pdfsubject={q-bio.NC, q-bio.QM},
pdfauthor={David S.~Hippocampus, Elias D.~Striatum},
pdfkeywords={First keyword, Second keyword, More},
}

\begin{document}
\maketitle

\begin{abstract}
	We propose a general framework for the Discontinuous Galerkin-induced Neural Network (DGNN), inspired by the Interior Penalty Discontinuous Galerkin Method (IPDGM). 
    In this approach, the trial space consists of piecewise neural network space defined over the computational domain, while the test function space is composed of piecewise polynomials. 
    We demonstrate the advantages of DGNN in terms of accuracy and training efficiency across several numerical examples, including stationary and time-dependent problems. 
    Specifically, DGNN easily handles high perturbations, discontinuous solutions, and complex geometric domains.
\end{abstract}


\section{Introduction}

Partial Differential Equations (PDEs) are fundamental to the mathematical modeling of a wide range of physical, biological, and engineering systems. 
These systems span diverse fields, such as fluid dynamics, heat transfer, structural mechanics, and population dynamics. 
Traditionally, numerical methods such as Finite Difference Method (FDM), Finite Volume Method (FVM), Finite Element Method (FEM) \citep{FDM, FEM}, and spectral methods have been employed to obtain approximate solutions to PDEs. 
Although these methods are highly effective, they ultimately reduce to solving linear systems of equations. This often leads to several challenges, such as when solving high-dimensional problems or performing adaptive solutions for complex domains. In such cases, the stiffness matrix can become exceedingly large and sparse.
Therefore, as the conditions of the PDE become more complex, these traditional numerical methods become computationally expensive. Moreover, classical discretization methods heavily depend on the resolution of the grid partition. This implies that higher accuracy requires finer grid partitions, which also leads to larger stiffness matrices. Meanwhile, we are limited to obtaining values only at the grid points, called \textbf{Resolution-Variance}.

Notably, deep learning methods have gained increasing attention in recent years. It has been successfully applied across various fields. In particular, in areas such as computer vision \citep{voulodimos2018deep, inbook}, natural language processing \citep{minaee2024largelanguagemodelssurvey}, and protein structure prediction \citep{SeniorAndrewW2020Ipsp}, deep learning has significantly improved both the efficiency and accuracy of problem-solving, owing to its powerful feature learning capabilities and data-driven advantages. 
In the field of scientific computing, deep learning, as a viable alternative or complement to traditional methods, has been employed to solve equations, such as ODEs \citep{DBLP:journals/corr/abs-1806-07366}, PDEs \citep{kharazmi2019variationalphysicsinformedneuralnetworks}, SODEs \citep{Zhu_2023, song2021scorebasedgenerativemodelingstochastic}, SPDEs \citep{DBLP:journals/corr/abs-2110-10249}, with application in fluid dynamics, climate prediction and structural mechanics, etc. 
Through end-to-end learning and optimization, deep learning models, like Neural Operators \citep{li2021fourierneuraloperatorparametric, Lu_2021, DBLP:journals/corr/abs-2010-03409}, can automatically extract valuable patterns and insights from large datasets, providing new perspectives and techniques that complement traditional numerical methods. 

Neural networks, particularly deep neural networks (DNNs), have demonstrated universal function approximation capabilities, making them well-suited for approximating functions. 
We refer to this model, which uses deep neural networks (DNNs) as surrogate solvers, as \textbf{DNN-based solvers}. 
Unlike the Resolution-Variance of classical numerical methods, deep learning approaches can directly learn continuous solutions from data or the governing equations, offering \textbf{Resolution-Invariant} solution in certain cases. 
These advantages are especially pronounced in high-dimensional problems or scenarios requiring real-time solutions, where traditional methods become prohibitively costly. 

During the past decades, DNN-based solvers have offered several novel methodologies that diverge from traditional numerical solvers. 
A particularly promising class of models includes Physics-Informed Neural Networks (PINNs) \citep{raissi2017physicsinformeddeeplearning}, which embed the governing PDEs, boundary conditions(BCs), and initial conditions(ICs) directly into the loss function. 
This allows the network to learn a solution that satisfies both the physical laws and the observed data. 
To deal with complex spatial domains, PINNs have been extended to include domain decomposition techniques \citep{Jagtap2020ExtendedPN, Moseley2023}, and domain mapping \citep{GAO2021110079}.
Another class of DNN-based solvers takes a different approach by formulating the loss function based on the weak or variational form of the PDEs, \citep{e2017deepritzmethoddeep,khodayimehr2019varnetvariationalneuralnetworks,Kharazmi_2021,Zang_2020,zang2023particlewnnnovelneuralnetworks,bao2024pfwnndeeplearningmethod,8918421}. 
To deal with the dimension time $t$ better, many novel frameworks are proposed, such as PI-LSTM \citep{zhang2020physics}. However, these methods are all autoregressive, leading to errors accumulation over time.
Recently, operator learning \citep{Lu_2021, Cai_2021, li2021fourierneuraloperatorparametric} is developed to learn mappings between function spaces. 
These approaches enable fast and accurate predictions of PDE solutions by leveraging high-fidelity data typically generated from classical numerical methods or the real world. 
However, acquiring such data can be challenging and resource-intensive.

It is important to emphasize that DNN-based and traditional numerical solvers are not two independent methods classes. 
Classical numerical methods are often designed from the perspective of approximating continuous problems through discretization. 
Their advantage lies in the flexibility to handle non-uniform grids by adaptive discretization. But they inevitably introduces truncation errors. 
In contrast, DNN-based solvers approximate solutions using continuous models, which eliminates truncation errors but comes with challenges in handling non-uniform grids and capturing local information effectively.
Many studies have attempted to integrate both approaches, such as \citep{kan, finitenet}. 
However, approaches such as using convolutional kernels to approximate finite difference methods neither resolve the issue of truncation errors nor effectively address the challenge of unstructured grids.
MeshGraphNet \citep{DBLP:journals/corr/abs-2010-03409} proposed the Graph Neural Networks to solve PDEs on unstructured grids. However, this approach is more analogous to image processing, often resulting in challenging model training and a large number of parameters.

Many of the challenges, like slow training speed and even training failures of DNN-based models, are typically attributed to the network's inability to capture nonlinearities effectively. 
Since a DNN model serves as the surrogate solver for the global solution, the training process often exhibits a phenomenon where some local loss changes can significantly impact the entire system.

In this paper, we introduce \textbf{DGNN}, a novel DNN-based framework for solving PDEs. 
Our approach effectively integrates the advantages of both traditional numerical methods and deep learning approaches. 
Utilizing fewer network parameters provides the flexibility to handle non-uniform grids and mitigates truncation errors in numerical solutions.
The approach is inspired by Discontinuous Galerkin Methods(DGM) \citep{DG,computedg}. DG methods are a class of finite element methods using completely discontinuous basis functions. 
This allows these methods to have some flexibility which is not shared by typical finite element methods, such as arbitrary triangulation, p adaptivity, parallel efficiency and local conservation.
The framework we proposed is similar to DG methods, but we introduce a DNN model for every single element, which serves as the trial function space \textbf{piecewise DNN function space} $\mathcal{N}_{\Omega_h}$.
The test function space is defined as a piecewise polynomial space that is not compactly supported, consistent with the DG method. 
This hybrid approach merges the interpretability and rigor of the weak formulation with the powerful function approximation capabilities of deep learning.

The main contributions of this paper include:
\begin{itemize}
    \item Novel DNN-based framework for PDE solving: We propose a new deep neural network framework that adopts a divide-and-conquer strategy for training on non-uniform grids. By designing parallel linear layers, our method achieves Resolution-Invariant and high parallelism.
    \item Innovative numerical flux-based communication: Departing from traditional parameter weight-sharing mechanisms, we introduce a novel numerical flux-based information exchange paradigm between networks. This weaker communication modality exhibits enhanced efficiency in solving complex PDE scenarios, including multi-scale solutions, irregular domain problems, and discontinuous solution challenges.
    \item Local refinement training algorithm: We implement a localized tuning algorithm during network training. Numerical experiments demonstrate that our method achieves more stable training dynamics, faster convergence rates, and better preservation of local conservation properties compared to conventional approaches.
\end{itemize}

The remainder of the paper is organized as follows.
The related works and classical DG method will be discussed in Section \ref{related work}.  
In Section \ref{DGNN}, the framework and algorithms of DGNN are proposed for solving PDEs. It includes the construction of weak forms and details of training for DGNN. 
In Section \ref{experiments}, the accuracy and efficiency of the proposed methods are examined through several numerical experiments. Finally, we give some conclusions and discussions in Section \ref{conclusion}. All relevant code has been made available and can be accessed at the following GitHub repository: \url{https://github.com/cgymmy/DGNN}

\section{Related Work\label{related work}}
Methods for solving PDEs can generally be categorized into three broad types: purely physics-driven, purely data-driven, and hybrid methods that integrate both physics and data-driven approaches \citep{Karniadakis2021}. 
The purely physics-driven methods usually refer to classical numerical methods, like FDM, FEM, and FVM. The problems they address typically involve differential equations with well-defined analytical forms. The purely data-driven methods refer to operator learning methods, like DeepONet \citep{Lu_2021}, FNO \citep{li2021fourierneuraloperatorparametric}, etc. Data-driven approaches are often applied to scenarios where mechanistic information is incomplete, like Multi-physics and Multiscale problems. These models generally have strong generalizability and robustness for their noisy data. In contrast, hybrid methods, such as physics-informed neural networks (PINNs) \citep{raissi2017physicsinformeddeeplearning}, are primarily designed for solving individual PDE systems with well-defined equations in a data-free setting or for addressing inverse problems when data is available. Here, we focus on the problems targeted by hybrid methods.

The hybrid methods can further be classified into two categories: strong-form methods and weak-form methods.
For instance, we consider PDE with the Dirichlet boundary condition of the form:
\begin{equation}\left\{
    \begin{aligned}
        &\mathcal{L}u(x, t) = f(x, t), \qquad (x, t)\in \Omega \times [0, T]\\
        &u(x, 0) = g(x),\qquad x\in \Omega\\
        &u(x, t) = h(x, t), \qquad x\in \partial \Omega
    \end{aligned}\right.
\end{equation}
where $\mathcal{L}$ is differential operator, $\Omega$ is the space domain.
\subsection{Strong Form Solver}
Strong-form methods focus on directly approximating the solution of PDEs by enforcing the residual of the governing equations over the domain and the boundary. 
This is typically done by minimizing the residuals at collocation points within the domain and on the boundaries.
The typical class of strong-form solvers is PINN and its variant. The loss function of PINN is the strong form of PDEs:
\begin{equation}
    \begin{aligned}
        \mathcal{R}(u_\theta)&=\sum_i^{N_p}\|\mathcal{L}u_\theta-f\|_2^2 + \sum_i^{N_b} \|u_\theta-h\|_2^2+\sum_i^{N_I} \|u_\theta-g\|_2^2\\
        &=\mathcal{R}_E + \mathcal{R}_B + \mathcal{R}_I
    \end{aligned}
\end{equation}
where $N_p, N_b,$ and $N_i$ represent the number of estimated points in the domain on the Dirichlet boundary and the initial boundary.  
Additionally, $u_\theta$ is a surrogate model based on a deep neural network (DNN), which takes $(x, t)$ as input and consists of $l$ hidden layers, each containing $N$ neurons, $\theta$ is the parameters of the network. 

For the exact solution, the residuals are identically zero. Then, the problem can be formulated as the minimization problem:
\begin{equation}
    u_\theta = \arg \min_{u_{\theta}}\mathcal{R}(u_\theta)
\end{equation}
PINNs have been further extended to address various types of PDEs, such as fractional advection-diffusion equations (fPINN) \citep{Pang_2019}, Navier-Stokes equations (NSFnets) \citep{Jin_2021}, etc. Several improvements on PINN, such as PDEs in the irregular domain (PhyGeoNet) \citep{GAO2021110079}, domain decomposition techniques (XPINN) \citep{Jagtap2020ExtendedPN}. Meanwhile, loss function modifications have been proposed to improve their efficiency and accuracy in solving complex PDEs.
Some training techniques include performing feature transformations on the input $(x, t)$ like the Fourier features $(x,t)\rightarrow (sin(x), cos(x), \cdots, sin(x)cos(x))$ in \citep{bao2024pfwnndeeplearningmethod}, imposing hard constraints.

As mentioned in \citep{Moseley2023}, whilst PINN performs well for low frequencies, it struggles to scale to higher frequencies; the higher frequency PINN requires many more free parameters, more collocation points, converges much more slowly, and has worse accuracy. Moseley solves this problem by defining some overlapping sub-domains, usually uniform meshes. However, Moseley's work establishes a window function that requires scanning the entire domain for local refinements. Although multi-threaded parallelization can accelerate this process, it imposes strict requirements on the discretized grid. For instance, a triangulated mesh cannot be easily partitioned using a fixed window function, making inter-thread communication challenging. Without multi-thread communication, the complexity of fBPINN increases linearly. However, in our proposed method, we completely circumvent this issue by defining parallel linear layers.

\subsection{Weak Form Solver}
Inspired by the weak form of PDEs, these methods leverage the variational or weak formulation of PDEs, which is more amenable to numerical optimization. 
Meanwhile, the order of differential operators can be effectively reduced by "integrating by parts," which reduces the required regularity in the (nonlinear) solution space and allows for more flexibility in approximating non-smooth or discontinuous solutions.
Compared to the strong form, when dealing with higher-order differential equations, a lower differentiation order leads to fewer automatic differentiation, significantly reducing the computational overhead.
We also find that weak-form methods naturally enable local learning through domain decomposition, facilitating more efficient and parallelizable training, particularly on complex or irregular domains. 
These features make weak-form methods particularly well-suited for deep learning frameworks to solve PDEs.

The VPINN \citep{khodayimehr2019varnetvariationalneuralnetworks} first developed a variational form of PINN within the Petrov-Galerkin framework. 
The variational residuals for stationary PDE are defined as:
\begin{equation}    
        \mathcal{R}_v=\|\int_{\Omega}\left(\mathcal{L}_x (u_{N N})-f\right) vd\Omega\|_2^2 + \mathcal{R}_B
\end{equation}
where $\mathcal{L}_x$ is differential operator for stationary equation, $v(x)$ is test function. Here in VPINN, the test function is chosen as Sine and Cose. 
Later, hp-VPINN \citep{Kharazmi_2021} is developed by taking a different set of localized test functions, defined over non-overlapping subdomains($\Omega = \cup_{i}^N E_i$). 
Building on VPINN, the Weak Adversarial Network (WAN) \citep{Zang_2020} and Variational Neural Networks (VarNet) \citep{khodayimehr2019varnetvariationalneuralnetworks} use operator norm minimization and Petrov-Galerkin principles, respectively, to improve the efficiency of solving PDEs.

Different choices of trial function $u_\theta(x,t)$ and test function $v(x,t)$ correspond to different methods as in Table(\ref{testfunction}). 
The Deep Ritz Method (DeepRitz) \citep{e2017deepritzmethoddeep} approximates the solution by minimizing the corresponding energy functional, a special case of weak form.
We can also take PINN as the special case of weak form with the test function as the Dirac Function. 
In WAN \citep{Zang_2020}, the test function is chosen to be a DNN, which serves as an adversarial network by defining a min-max form. 
In recent work PWNN \citep{zang2023particlewnnnovelneuralnetworks}, the test function is designed especially as a function with compact support, such as Wendland's function. 
This avoids the integration on the element boundary, but it leads to an incredible number of testing points and a special training strategy.

\begin{table}[H]
    \centering
    \begin{tblr}{hlines, vlines}
    \textbf{Solution} $u_\theta$                                       & \textbf{Test Function v}                           & \textbf{Test Domain}                \\
    PINN \citep{raissi2017physicsinformeddeeplearning}  & Delta Function $\delta$ & Collocation points         \\
    hPINN \citep{kharazmi2019variationalphysicsinformedneuralnetworks} & Sine and Cose                              & Whole domain               \\
    hp-PINN \citep{Kharazmi_2021}                     & Legendre Polynomials                              & Local area (Square domain) \\
    VarNet \citep{khodayimehr2019varnetvariationalneuralnetworks} & Nodal Basis(Polynomials) & Local area  \\
    WAN \citep{Zang_2020}                              & DNN                                      & Whole domain               \\
    PWNN \citep{zang2023particlewnnnovelneuralnetworks} & CSRBFs                                   & Local area (Random Sphere domain) 
    \end{tblr}
    \caption{Different choices of Test Function}
    \label{testfunction}
\end{table}

The main challenges highlighted in \citep{Kharazmi_2021} are truncation and numerical integration errors. 
The truncation error, caused by the increasing number of test functions, may lead to optimization failures. For example, in hp-VPINN, the order of test functions can be up to 60.
Meanwhile, when the test domain is chosen to be the entire domain, the numerical integration error arises due to an insufficient number of integration points over the whole domain. 
However, when the test region is selected as a local subdomain, training becomes highly unstable due to global parameter sharing. However, the "divide and conquer" strategy we propose effectively mitigates this issue.

\subsection{Discontinuous Galerkin Method}
The main idea of the Discontinuous Galerkin method \citep{DG} is similar to that of traditional finite element methods: selecting basis functions and ultimately solving a linear equation. 
However, the key difference is that DG methods allow for discontinuities in the solution by choosing completely discontinuous basis functions. 
This makes each element independent of its neighbors.
There are various approaches within DGMs, such as interior penalty methods, local DG methods, and hybrid DG methods. 
Here, we focus on the Interior Penalty Discontinuous Galerkin Method(IP-DGM) \citep{computedg}. For time-dependent PDE, DG takes high-order time discretization as FEM. Hence, for simplicity, we use Poisson Equation with Dirichlet boundary condition as an example: 
\begin{equation}\label{Poisson}
    \left\{\begin{aligned}
    &-\Delta u=f\qquad x \in \Omega\\
    &u=g \qquad x\in \partial \Omega\\
\end{aligned}\right.\end{equation}
where $\Omega$ is the domain, $\partial \Omega$ is the Dirichlet boundary. 
The domain $\Omega$ can be discretized into a mesh $\Omega_h = \cup_{i=1}^N E_i$ through triangulation, where $E_i$ denotes the individual triangular elements of the mesh. 
Denote the set of all edges in $\Omega_h$ as $\mathcal{E}_h$, which can be decomposed into the set of interior edges $\mathcal{E}_h^I$ and Dirichlet boundary edges $\mathcal{E}_h^D$. 
Thus, we have the relationship $\mathcal{E}_h=\mathcal{E}_h^I\cup\mathcal{E}_h^D$. 
By taking test functions in a finite element space $V_h^k$ consisting of piecewise polynomials:
\begin{equation}
    V_h^k = \{v:v|_{E_i}\in \mathbb{P}_k(E_i),\forall E_i\in \Omega_h\}
\end{equation}
where $\mathbb{P}_k(E_i)$ denotes the set of piecewise polynomials of degree up to $k$ defined on the cell $E_i$. Then the weak form of the Poisson Equation can be written as: 
find the unique function $u\in V_h^k$ s.t. $\forall v\in V_h^k$:
\begin{equation}\label{weak form}
    \begin{aligned}
    \int_\Omega \nabla u\cdot \nabla v - \int_{\partial \Omega}(\nabla u \cdot \textbf{n})v=\int_\Omega fv
    \end{aligned}
\end{equation}
Here in DG method, the weak form, after dicretization, can be reformulated as: find a solution $u_h\in V_h^k$, s.t. $\forall v\in V_h^k$: 
\begin{equation}
\int_{E_i} \nabla u\cdot \nabla v - \int_{\partial E_i}(\widehat{\nabla u \cdot \textbf{n}})v-\int_{E_i} fv=0,\qquad \forall E_i\in \Omega_h
\label{weak}
\end{equation}
where $\widehat{F(u)}$ is defined as \textbf{Numeric Flux}:
\begin{equation}
    \begin{aligned}
        \widehat{F(u)} &= \frac{1}{2}(F(u)|_l + F(u)|_r) - (u|_l - u|_r)\\
        &= \left\{F(u)\right\} - \left[u\right]
    \end{aligned}
\end{equation}
Here, $\{\cdot\},[\cdot]$ are called \textbf{Jump} and \textbf{Average} operators, respectively. 
And for boundary edge $e \in \mathcal{E}_h^D$, we set $[u(x)]=\{u(x)\} = g(x),\forall x\in e$.
Here we introduce two notations:
\begin{equation}
    (f, g)_E = \int_E fg\qquad \langle f, g\rangle_e = \int_e ([f]\{g\} + \{f\}[g]) 
\end{equation}
where E is the element in the mesh $\Omega_h$, and e is the edge. 
Hence, the weak form can be rewritten as
\begin{equation}\label{1-3}
    (\nabla u, \nabla v)_{\Omega_{h}}-\langle\{\nabla u \cdot \textbf{n}\},[v]\rangle_{\mathcal{E}_{h}^{I D}}=(f, v)_{\Omega_h}, \quad \forall v \in V .
\end{equation}
Then, we can define the bilinear form:
\begin{equation}
    \left\{\begin{aligned}
    &A_{h}(u, v):=  (\nabla u, \nabla v)_{\Omega_{h}}-\left(\langle\{\textbf{n}\cdot \nabla u \},[v]\rangle_{\mathcal{E}_{h}^{I D}}-\epsilon\langle\{\textbf{n}\cdot\nabla v \}, [u]\rangle_{\mathcal{E}_{h}^{I D}}\right)+J_{0}(u, v)+J_{1}(u, v)\\
    &F_h(v)=(f, v)_{\Omega_h}+\epsilon\langle\{\textbf{n}\cdot\nabla v \}, [u]\rangle_{\mathcal{E}_{h}^{D}} + \sum_{e \in \mathcal{E}_{h}^{D}} \frac{\sigma_{0}}{h_{e}} \int_{e}g_Dv\\
\end{aligned}\right., \qquad \forall v \in V
\end{equation}
where the penalty terms are defined as:
$$J_{0}(u, v):= \sum_{e \in \mathcal{E}_{h}^{I D}} \frac{\sigma_{0}}{h_{e}} \int_{e}[u][v],\qquad
J_{1}(u, v):= \sum_{e \in \mathcal{E}_{h}^{I}} \sigma_{1} h_{e} \int_{e}[\nabla u \cdot n][a \nabla v \cdot n]$$

Then we need to seek $ u_{h} \in V_h^k$ s.t.
\begin{equation}\label{dgPoisson}
    A_{h}\left(u_{h}, v_{h}\right)=F_h(v_h) \quad \forall v_{h} \in V_h^k.
\end{equation}
Hence, if we set basis functions ${\phi_i(x)}$ for $V_h^k$ s.t.
$u_h=\sum_{i=1}^{N_h}u_i\phi_i$. Then we have the form
$$\sum_{j=1}^{N_h}A_h(\phi_j,\phi_i)u_j=F_h(\phi_i),i=1\cdots N_h$$
which can be rewritten as $AU=F$. By solving this linear system, we can get the solution of PDEs on grid points.

\section{Discontinuous Galerkin Neural Network\label{DGNN}}
In this section, the framework of the Discontinuous Galerkin Neural Network (DGNN) is introduced. In DGNN, the loss function is based on the weak forms of the models, and the solution is parameterized as several deep neural networks.
Similar to the classical DG method, the test function space is defined as a piecewise polynomial space, while the trial function space is defined as \textbf{piecewise neural network space}.
\subsection{The Framework of DGNN}
\begin{figure}
    \centering
    \includegraphics[width=\linewidth]{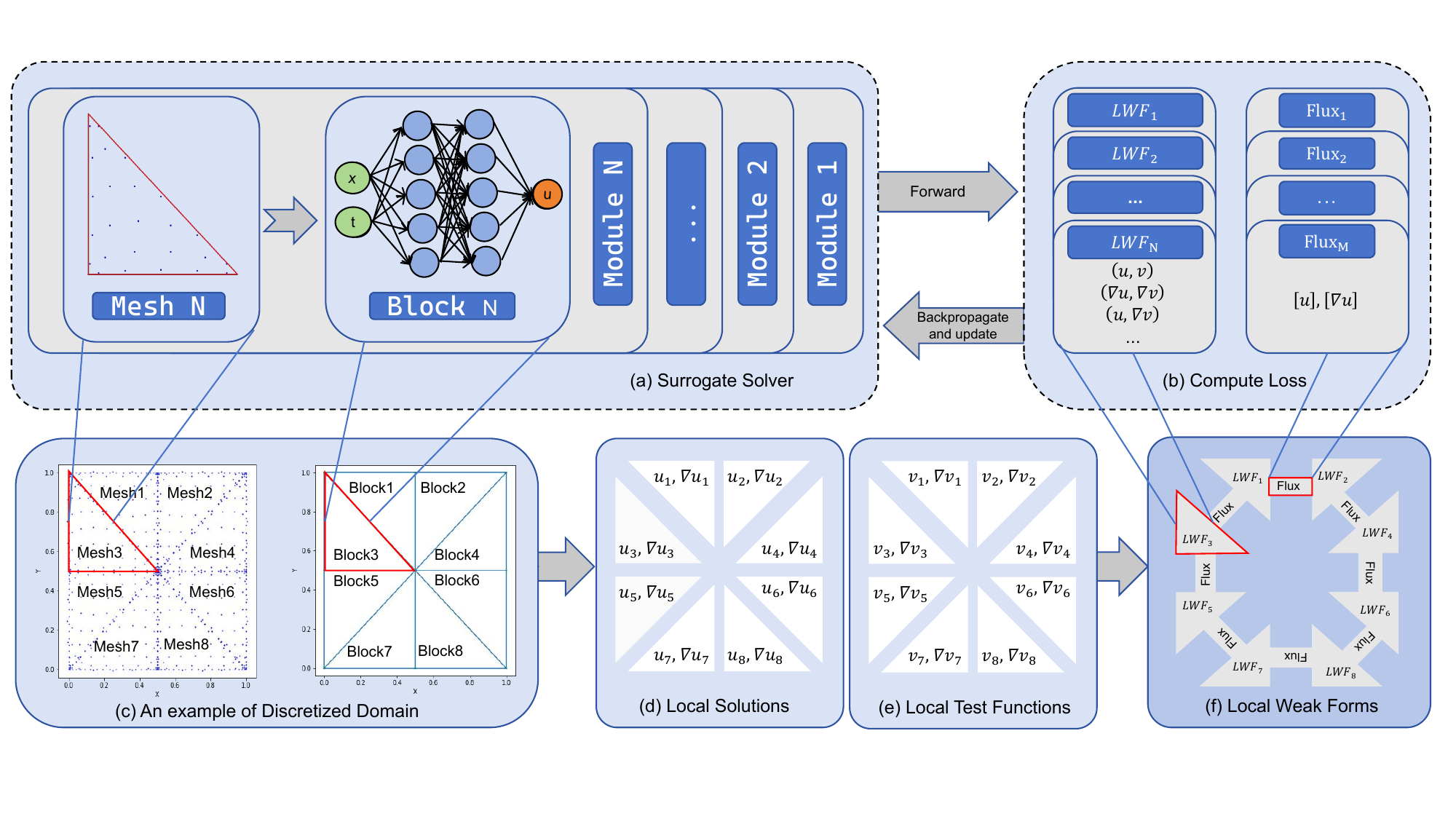}
    \caption{The sketch structure of DGNN. (a) The whole surrogate solver. The surrogate solver consists of N modules. Each module has a domain mesh and a block network. (b) Components of Loss Function. With N modules, we can compute N local weak-form equations and corresponding Numerical Flux. (c) An example of a discretized domain. (d)(e) The corresponding local solution and test functions. (f) The local weak forms and numerical flux.}
    \label{nn structure}
\end{figure}

Consider the general PDE of the form:
\begin{equation}\label{CDE}
    \left\{
    \begin{aligned}
        &u_t+\nabla \cdot F(u) = \nabla\cdot (D\nabla u) + f \qquad (x, t)\in \Omega\times [0, T]\\
        & u(x, t) = g(x) \qquad x\in \partial \Omega\\
        & u(x, 0) = u_0 \qquad x\in \Omega
    \end{aligned}
    \right.
\end{equation}
where $\Omega$ is the spatial domain, $\nabla \cdot F(u)$ is the convection term, and $\nabla\cdot(D\nabla u)$ is diffusion term, $D$ is the coefficient, $f$ is the source term. Then we have the weak form of Equ (\ref{CDE}): find $u\in L^2((0, T), H^1(\Omega))$ s.t. $\forall v\in H^1(\Omega)$
\begin{equation}\label{Weak CDE}
    \int_0^T\int_\Omega \left[u_tv+D\nabla u\cdot\nabla v-\nabla v\cdot F(u)\right]d\Omega dt - \int_0^T\int_{\partial \Omega}\left((D\nabla u-F(u)\cdot n\right)vdsdt = \int_0^T\int_\Omega fvd\Omega dt
\end{equation}
With appropriate boundary conditions and source functions, the weak form of (\ref{Weak CDE}) has a unique solution $u$. Typically, DNN-based methods approximate the solution \( u \) by a single model, i.e., using a single neural network \( u_{\text{NN}} \) as the surrogate solution. Such neural network usually consists of $l(l\geq 3)$ hidden layers with $n$ neurons per layer and activation function $\sigma$. Here we note the function space of nerual network as $\mathcal{N}_{l,n}=\{u_\theta(x,t|l, n):x\in \Omega, t\in [0, T], l,n\in \mathbb{N}^+\}$. 
Same as the DG method, we have domain discretization $\Omega_h=\cup_i^N E_i$. Then we define the \textbf{Piece-WIse DNN Space} $\mathcal{N}_{\Omega_h}$ as:
\begin{equation}
    \mathcal{N}_{\Omega_h}=\{u|_{E_i}(x, t)\in \mathcal{N}_{l,n}(E_i),\forall E_i\in \Omega_h\}
\end{equation}
where $\mathcal{N}_{l,n}(E_i)$ is space of neural networks having support $E_i$. 
This implies that each element is assigned a separate DNN-based model serving as the surrogate solution. 
Each of these models is constructed using only shallow neural networks with $l \leq 2$. 
Thus, the weak solution $u_\theta$ in DGNN takes the following form:
\begin{equation}
        u_\theta(x, t;\theta) = \sum_{i=1}^{N_h}u^i_{NN}(x, t; \theta_i),\qquad
        u^i_{NN} = \left\{\begin{aligned}
            &\sigma^{(l)} \circ \sigma^{(l-1)} \circ \cdots \circ \sigma^{(1)}(x, t),&\quad (x, t)\in E_i\times [0, T]\\
            &0, &\quad \operatorname{Otherwise}
        \end{aligned}\right.
\end{equation}
We believe that by divide and conquer", shallow neural networks can efficiently learn the behavior of the local solution within small local elements. 
Hence, for discretized mesh, the DNN-based weak form involves finding the solution $u_\theta\in \mathcal{N}_{\Omega_h}$, s.t. $\forall v\in V_h^k$:
\begin{equation}\label{nnweak CDE}
    \begin{aligned}
        &\int_0^T\int_{\Omega_h} \left[(\partial_t u_\theta)v+D\nabla u_\theta\cdot\nabla v-\nabla v\cdot F(u_\theta)-fv\right]d\Omega dt \\
        = &\int_0^T\int_{\partial \Omega_h}\left((D\nabla u_\theta-F(u_\theta)\cdot n\right)vdsdt 
    \end{aligned}
\end{equation}
Figure \ref{nn structure} illustrates the idea of DGNN with a sketch. 
First, the domain is triangulated into a non-uniform mesh shown in Figure \ref{nn structure}(c). It is worth noting that various discretization methods are available, including rectangle adaptive and triangular adaptive mesh discretization. Since triangulated meshes offer greater flexibility in handling irregular boundaries, we adopt Delaunay triangulation in this paper. Next, separate neural networks are placed within each element. To evaluate the integration of weak form, Gaussian-Legendre integration points are generated in each element. These mesh points and local neural networks form localized modules, and there are N modules in total(N is the number of elements). These N modules collectively constitute the entire surrogate solver, as illustrated in Figure \ref{nn structure}(a). Then, we can compute the local weak forms and Fluxes between elements like Figure \ref{nn structure}(b). Finally, we can compute the loss, backpropagate, and update.

\subsection{Loss Functions}
According to the DNN-based weak form (\ref{nnweak CDE}), we denote $\mathcal{R}$ as the residual of the weak form. It can be rewritten as:
\begin{equation}
    \begin{aligned}
        \mathcal{R} = &\sum_{i=1}^N\left[\int_0^T\int_{E_i} \left[(\partial_t u_\theta)v+D\nabla u_\theta\cdot\nabla v-\nabla v\cdot F(u_\theta)-fv\right]d\Omega dt \right.\\
        &\left.-\int_0^T\int_{\partial E_i}\left((D\nabla u_\theta-F(u_\theta)\cdot n\right)vdsdt\right] \\
        =& \sum_{i=1}^N\int_0^T \mathcal{S}_t^idt =\sum_{i=1}^N\mathcal{R}_i
    \end{aligned}
\end{equation}
where $\mathcal{S}_t^i$ and $\mathcal{R}_i$ are local semi and total weak form residual on $E_i$. Actually, for exact solution $\mathcal{S}_t^i$ is exact zero.
For the sake of simplicity, we study the $\mathcal{S}_t$ on a single element $E$. 

In Finite element methods, test functions are often taken as piecewise polynomials having compact support to avoid the integration on the element boundary. 
However, as in DG, the test functions are taken as piecewise polynomials, which allows for discontinuities. This provides the flexibility to handle non-smooth solutions and p adaptivity but takes the challenge of boundary integration.
However, it is not a big deal for scientific computing by CUDA, which will be discussed later. 
So here we take the test function $v\in V_h^k$ same as DG methods. For $\forall v\in V_h^k$, $v$ can be written as: 
\begin{equation}
    v = \left\{\begin{aligned}
        &\sum_{i=1}^{k}p_i(x),\qquad &x\in E\\
        &0, \qquad &x\notin E
    \end{aligned}\right., \qquad p_i(x)\in \mathbb{P}_i(E)
\end{equation}
where $P_i(x)$ is polynomial space of order $i$, having support $E$. For example, for 1D problem, take $v=\sum_i^n x^i$, while for 2D problems, take $v = \sum_i^n\sum_j^i x^jy^{i-j}$. 
Then $\nabla v=\sum_{i=0}^{k} \nabla p_{i}(x)$. We have:
\begin{equation}
    \begin{aligned}
        \|\mathcal{S}_t\|_2 \leq & \sum_{i=0}^{k}\left\|\int_{E}\left[(\partial_t u_\theta)p_i+D\nabla u_\theta\cdot\nabla p_i-\nabla p_i\cdot F(u_\theta)-fp_i\right]d\Omega\right.\\
        &\left.-\int_{\partial E}\left((D\nabla u_\theta-F(u_\theta)\cdot n\right)p_ids\right\|_2
    \end{aligned}
\end{equation}
Therefore, the local loss function for element $E$ is:
\begin{equation}
    \begin{aligned}
        \mathcal{L}_E(t) =& \sum_{i=0}^{k}\left\||E|\sum_{j}^{N_E}w_{E}^j\left[\partial_t u_\theta(x_j, t)p_i(x_j)+D\nabla u_\theta(x_j, t)\cdot\nabla p_i(x_j)-\nabla p_i(x_j)\cdot F(u_\theta(x_j, t))-f(x_j, t)p_i(x_j)\right] \right.\\
        &\left.-|\partial E|\sum_{j}^{N_e}w_{\partial E}^j\left((D\nabla u_\theta(x_j, t)-F(u_\theta(x_j, t))\cdot n_j\right)p_i(x_j)\right\|^2
    \end{aligned}
\end{equation}
where $N_E$ and $N_e$ are number of integration estimation point in element $E$ and on its boundary $\partial \Omega$. Here the method of numerical integration is chosen as Gauss-Legendre quadrature, where $w_E$ and $w_{\partial E}$ are weights of square and length integration. 
It should be noted that $N_E$ and $N_e$ are very small, usually $10-20$, as the discretized element is small enough. This explains why the challenge of boundary integration is efficiently solved. 
Hence, the global loss function is:
\begin{equation}
    \mathcal{L}_{eq} = \sum_{E_i\in \Omega_h}\sum_{j}^{N_t}\mathcal{L}_{E_i}(t_j)
\end{equation}
where $N_t$ is the number of time steps. In addition, to guarantee continuity, we need penalty terms:
\begin{equation}
    \begin{aligned}
        \mathcal{L}_{penalty} = &\sum_{e\in \mathcal{E}}\int_e\left( \left\|[u_\theta]\right\|^2 + \left\|[\nabla u_\theta]\right\|^2\right)\\
        \approx & \sum_{e\in \mathcal{E}} \sum_{i}^{N_{int}}\left( \left\|[u_\theta]\right\|^2 + \left\|[\nabla u_\theta]\right\|^2\right)
    \end{aligned}
\end{equation}

The DNN-based weak solution $u_\theta$ needs to satisfy the boundary condition, which has been contained in penalty terms, and initial condition:
\begin{equation}
    \mathcal{L}_{ic} = \sum_{i=1}^{N_i}\left\|u_\theta(x_i, 0)-u_0(x_i))\right\|^2
\end{equation}
where $N_i$ is the number of estimation points on each edge.
Finally, we can formulate our loss function as:
\begin{equation}
    Loss = \sigma_0\mathcal{L}_{eq} + \sigma_1\mathcal{L}_{ic} + \sigma_2\mathcal{L}_{penalty}
\end{equation}
where $\sigma_0, \sigma_1, \sigma_2$ are penalty coefficients. Then, the framework can be formulated as follows: We need to seek:
\begin{equation}
    u_\theta =  \arg_{u_\theta\in \mathcal{N}_{\Omega_h}}\min(Loss)
\end{equation}
\subsection{Algorithm and Details\label{details}}
The algorithm of DGNN for 1D problem is summarized in Algorithm \ref{alg1} and for 2D problem is summarized in Algorithm \ref{alg2}. But there are still some details that need to be clarified.

\begin{algorithm}[H]
    \begin{algorithmic}
        \State \textbf{Precomputing:} Partition Grids: $x$, number of sub-intervals $N$, integration mesh: 
        $x_{mesh}$, (gradience of) test function of order $p$ on reference mesh: $v, dv$.
        \State \textbf{Input:} $N, x, x_{mesh}, v, dv, \sigma$, num\_layers: $l$, hidden\_size: $N_{hidden}$, activation: act, max training iterations: $\operatorname{MaxIters}$
        \State \textbf{Initialize:} Network architecture $u_\theta(x, t)$, where $\theta$ indicates network parameters.
        \While{iter $<\operatorname{MaxIters}$}
            \State Generate $u_\theta(x_{mesh}, t)$ and $\nabla u_\theta(x_{mesh}, t)$ by auto-grad. 
            \State Separate the values of mesh points and grid points.
            \State Calculate $N$ local loss, and choose top K local losses
            \State Calculate the total loss.
            \State Backpropagate. Update the network parameters with Adam/LFBGS optimizer.
            \State iter +=1
        \EndWhile
        \State \textbf{Output:} The weak solution model $u_\theta$, which has $N$ modules corresponding to $N$ elements.
    \end{algorithmic}
    \caption{DGNN Algorithm For 1D Problem}
    \label{alg1}
\end{algorithm}

\begin{algorithm}[H]
    \begin{algorithmic}
        \State \textbf{Precomputing:} Triangulated Mesh $\Omega_h$, $\operatorname{x\_Mesh}$, test function: $v, \nabla_x v$, $w_{elt},w_{edge}$
        \State \textbf{Input:} $N, N_t, N_E, N_e, \sigma_i(i=1,2,3)$, num\_layers: $l$, hidden\_size: $N_{hidden}$, activation: act, max training iterations: $\operatorname{MaxIters}$
        \State \textbf{Initialize:} Network architecture $u_\theta(x, t)$, where $\theta$ indicates network parameters.
        \While{iter $<\operatorname{MaxIters}$}
            \State Generate $u_\theta(\operatorname{x\_Mesh}, t)$ and $\nabla u_\theta(\operatorname{x\_Mesh}, t)$ by auto-grad. \State Separate the value of inner points and boundary points.
            \State Calculate $N$ local loss, and choose top K local losses
            \State Calculate the total loss.
            \State Backpropagate. Update the network parameters with Adam/LFBGS optimizer.
            \State iter +=1
        \EndWhile
        \State \textbf{Output:} The weak solution model $u_\theta$, which has $N$ modules corresponding to $N$ elements.
    \end{algorithmic}
    \caption{DGNN Algorithm For 2D Problem}
    \label{alg2}
\end{algorithm}

\begin{figure}[ht]
    \centering
    \begin{minipage}[t]{0.3\textwidth}  
    \centering
    \includegraphics[width=\textwidth]{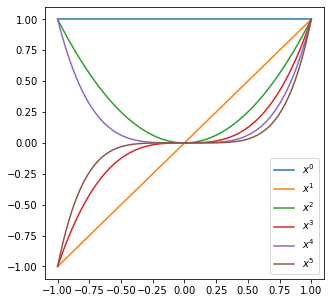}  
    \caption*{(a) 1D Test Functions}
  \end{minipage}
  \begin{minipage}[t]{0.65\textwidth}  
    \centering
    \includegraphics[width=\textwidth]{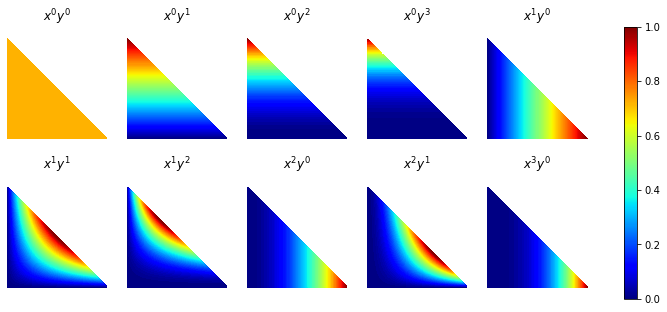} 
    \caption*{(b) 2D Test Functions}
  \end{minipage}
    \caption{Test Functions. (a) 1D test functions are chosen as 1D polynomials. The picture illustrates a one-dimensional polynomial of up to the fifth order. (b) 2D test functions are chosen as 2D polynomials. The displayed 2D polynomials have a maximum order of three.}
    \label{test functions}
\end{figure}

\begin{figure}[H]
    \centering
    \includegraphics[width=\linewidth]{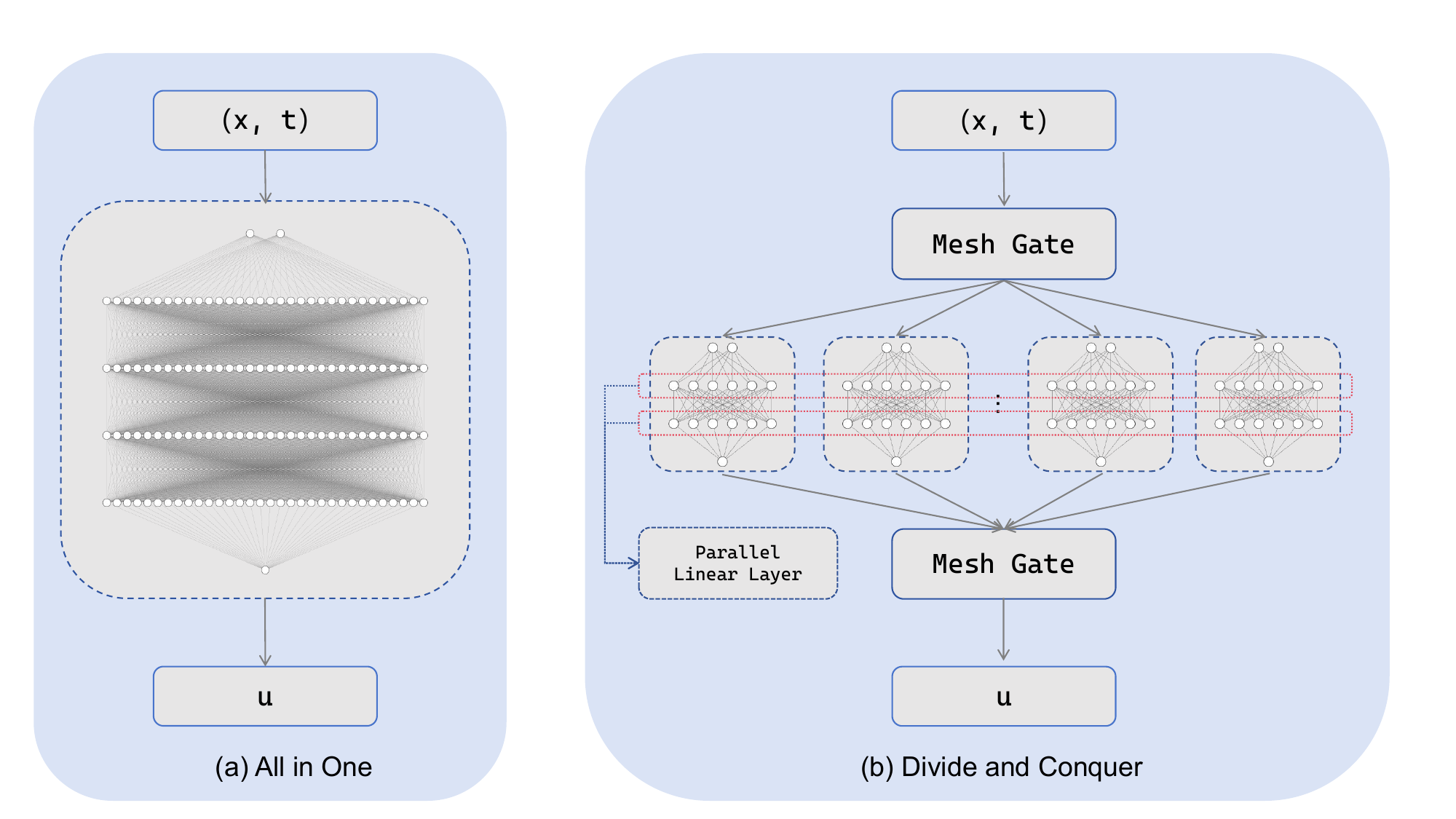}
    \caption{Net structure. The left column (a) is the "All in One" solver; The right column (b) is the "Divide and Conquer" solver we proposed. Here, the mesh gate is used to decide which block the estimate point belongs to. Actually, during the training process, the mesh gate is virtualized. And the red-boxed area is a scratch of the parallel linear layer.}
    \label{Net Structure}
\end{figure}

\paragraph{Test Functions: Fewer.} Take 2D cases as an example. Discretization is typically achieved through triangulation, such as Delaunay triangulation. Therefore, each discretized element is a triangle, ultimately forming a non-uniform mesh.
Since we define basis function $v\in V_h^k$ on reference triangular element $\hat{E}$, whose vertices are $(0,0),(1,0),(0,1)$. Hence for any triangular element, whose vertices are $(x_1, y_1),(x_2, y_2),(x_3, y_3)$, we have Affine transformation $F_E:\hat{E}\rightarrow E$:
$$\begin{pmatrix}
    x\\y
\end{pmatrix} = F_E\begin{pmatrix}
    \hat{x}\\ 
    \hat{y}
\end{pmatrix}=\begin{pmatrix}
    x_2-x_1 & x_3-x_1\\
    y_2-y_1 & y_3-y_1
\end{pmatrix}\begin{pmatrix}
    \hat{x}\\ 
    \hat{y}
\end{pmatrix}+\begin{pmatrix}
    x_1\\y_1
\end{pmatrix}=B_E\begin{pmatrix}
    \hat{x}\\ 
    \hat{y}
\end{pmatrix}+b_E$$ 
Then, the test function defined on the element has: 
\begin{equation}
    v(x)=\hat{v}(\hat{x}) = \hat{v}(F_E^{-1}(x)),\qquad\nabla_x v = B_E^{-T}\nabla_{\hat{x}} \hat{v}
\end{equation}
where $\hat{x}$ is points in reference triangle. These points are quadrature points in square integration, noted as $\operatorname{refp\_elt}$. 
Similarly, since we need to do integration on element boundary, Gauss-Legendre quadrature points are also generated on edges of the reference triangle, noted as $\operatorname{refp\_edges}$. By assembling $\operatorname{refp\_elt}$ and $\operatorname{refp\_edges}$, we have the reference mesh points $\operatorname{Ref\_Mesh}$ on reference triangle. According to Affine mapping, we can have mesh points $\operatorname{x\_Mesh}$ of the whole domain.
We should note that both the test function and its gradient can be precomputed in advance. For 1D cases, Affine mapping is similar, and corresponding elements are intervals. Test functions are shown in Fig (\ref{test functions})
\begin{remark}
    Since we define the neural network on local elements, the model gains more nonlinear fitting capability. Hence to approximate the weak solution, we don't require a large number of test functions. For instance, in our DGNN framework, the order of test functions $deg\leq 5$, even $deg=1$ has a good performance, while in hp-VPINN \citep{Kharazmi_2021}, they employ the order of  Legendre polynomials up to 60  as test functions.
\end{remark}
\paragraph{Test domains: Fewer and more flexible.}
Taking the 2D case as an example, the test domains are determined by the triangulation of the domain. Here, we use Delaunay triangulation for discretization. Naturally, it can be extended to adaptive discretization. Due to the flexibility of triangulation, DGNN can adapt to various complex boundaries while reducing the test region.
\begin{remark}
    In PWNN \citep{zang2023particlewnnnovelneuralnetworks}, $N_p$ test particle domains are randomly generated randomly during each training epoch. For instance, at least $N_p= 100$ test particle domains, where 10 integration points are generated, should be chosen in each iteration for 1D Poisson$\omega = 15\pi$(Summing up to 1000 points). While in the framework we proposed, $N=10, N_{int}=20$(Summing up to 500 points) is enough. Fewer test domains mean fewer integrations. 
\end{remark}
\paragraph{Training: Fewer parameters and higher parallelism.}
In fact, the idea of divide and conquer originates from the classical FEM. Here, by defining parallel linear layers, we fully leverage the efficient matrix computation capabilities of neural networks. In other words, we transform a deep neural network of a certain width into a relatively wide but shallow neural network as a surrogate solver, as shown in Figure \ref{Net Structure}.

Moreover, in this framework, the model parameters are separated rather than directly shared. Instead, information is transmitted between parameters through Numerical Flux, which serves as a \textbf{weaker form of sharing}. The advantage of this strategy is that it enables more efficient and stable training, preventing local errors from affecting global parameters. As a result, the proposed framework is able to maintain local conservation properties.

\section{Experiments\label{experiments}}
In this section, we present a series of numerical experiments to demonstrate the effectiveness of the DGNN method in overcoming key challenges, including the need for a large number of integration points in the weak form, adaptability to irregular domains, and the ability to handle multiscale and discontinuous solutions. We compare our proposed method with some widely used DNN-based models: vanilla PINN and its weak form variant hp-VPINN, which is VPINN when its number of sub-domains equals 1, DeepRitz Method, and PWNN. 
\paragraph{Experimental setups} Unless otherwise specified, both the PINN-type and PWNN models employ multilayer perceptrons (MLPs), whereas the DeepRitz method utilizes a residual neural network (ResNet). All methods use the Adam optimizer with a learning rate of \(10^{-3}\). In DGNN, we employ the L-BFGS optimizer together with the Adam optimizer at a learning rate of \(10^{-4}\). With our framework, the penalty coefficients are set as $\sigma_i=1$.
All related experiments were executed using GeForce RTX 3090.

\subsection{1D Poisson Equation}
We first consider the 1D Poisson Equation with high-frequency perturbation of the form:
\begin{equation}\label{1dPoisson}
    \left\{
    \begin{aligned}
        &-\frac{\partial^2u}{\partial x^2}(x) = f(x) \qquad x\in \Omega=(0,\frac{3}{2})\\
        &u(x)=g(x)\qquad x\in \partial \Omega
    \end{aligned}
    \right.
\end{equation}
We construct the equation by the real solution $u(x)=xcos(\omega x)$, having corresponding source term $f(x)$ and boundary condition $g(x)$. We use DGNN to approximate the target function. This case originated from PWNN and fBPINN. \citep{basir2022criticalinvestigationfailuremodes, Moseley2023} also stated that vanilla PINN struggles to solve solutions with high frequency. 

In this work, we consider two scenarios for the Poisson equation: one featuring a low-frequency solution, $\omega=3\pi$ and the other featuring a high-frequency solution, $\omega=15\pi$. For the low-frequency $\omega=3\pi$ solution, the corresponding parameter settings are as follows: As for our approach DGNN, we set $N=5, N_{int}=20, p=5$ with DGNN $(l=2, N_{hidden}=40, act=tanh)$. For vanilla PINN, 10000 collocation points are chosen in $\Omega$, and the model is MLP($l=3, N_{hidden} = 40$). For the DeepRitz method, we generate 200 Gauss-Legendre Quadrature points in $\Omega$, and the model is ResNet($l=6, N_{hidden} = 40$, act=tanh). Also, the boundary penalty coefficient is 500. In hp-PINN, the domain $\Omega$ is divided into $N=5$ sub-domains, and $N_{int}=30$ on each sub-interval. The hp-VPINN model is MLP($l=6, N_{hidden}=50, act=tanh$) and $p=6$. Other parameters maintain consistent settings: The Optimizer is Adam with a learning rate of $1e-3$. For the high-frequency $\omega=15\pi$ solution: Networks of PINNs, hpVPINN, and DeepRitz, are set to be deeper and wider: For PINN, the model is MLP($l=5, N_{hidden} = 128$). But for DGNN, the only change is the number of sub-intervals $N=25$.

\begin{figure}[ht]
    \centering
    \begin{minipage}[t]{0.45\textwidth}  
    \centering
    \includegraphics[width=\textwidth]{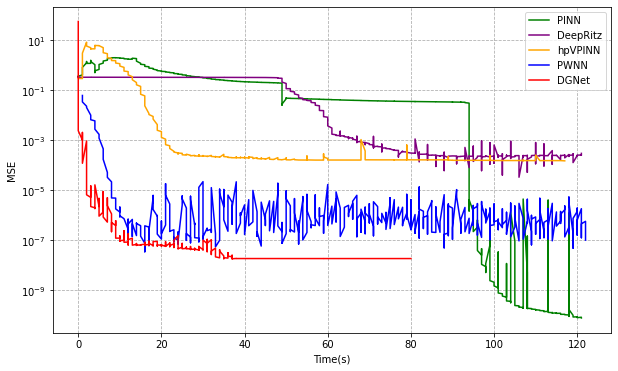}  
    \caption*{MSE vs. Time}
  \end{minipage}
  \begin{minipage}[t]{0.45\textwidth}  
    \centering
    \includegraphics[width=\textwidth]{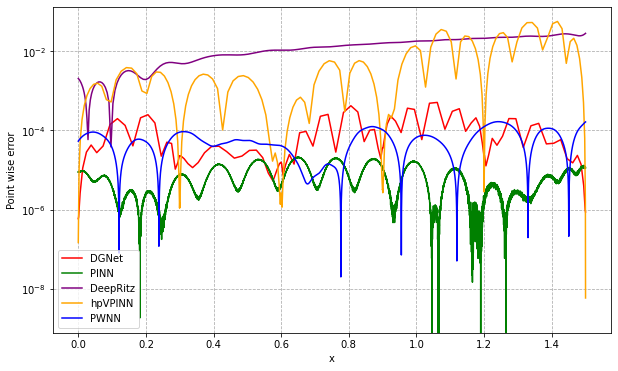} 
    \caption*{Pointwise Error}
  \end{minipage}
    \caption{The performance of different methods in solving 1D Poisson \ref{1dPoisson} with $\omega =3\pi$.}
    \label{1dPoisson3pi}
\end{figure}

\begin{figure}[H]
    \centering
    \begin{minipage}[t]{0.45\textwidth}  
    \centering
    \includegraphics[width=\textwidth]{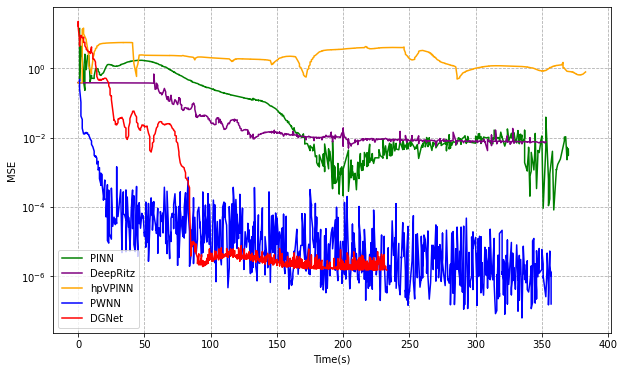}  
    \caption*{MSE vs. Time}
  \end{minipage}
  \begin{minipage}[t]{0.45\textwidth}  
    \centering
    \includegraphics[width=\textwidth]{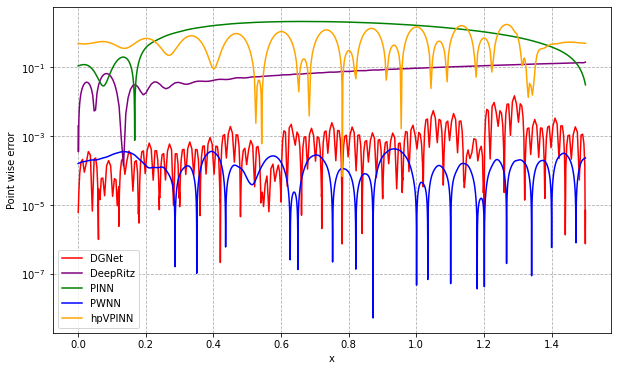} 
    \caption*{Pointwise Error}
  \end{minipage}
    \caption{The performance of different methods in solving 1D Poisson \ref{1dPoisson} with $\omega = 15\pi$.}
    \label{1dPoisson15pi}
\end{figure}

The result is illustrated in Figure \ref{1dPoisson3pi} and Figure \ref{1dPoisson15pi}. We observe that PINN, DeepRitz, and hpVPINN converge more slowly and are prone to local error issues compared to PWNN and DGNN. Moreover, DGNN achieves faster convergence than PWNN and attains higher accuracy (in terms of MSE). During the later stages of PWNN training, the MSE notably fluctuates, whereas DGNN maintains a stable error profile. This discrepancy arises because classical algorithms, such as PINN, hpVPINN, PWNN, etc., employ a single, global DNN-based solution. This makes the model susceptible to local loss deficiencies that can propagate and adversely affect the entire parameter set $\theta$. In contrast, DGNN adopts a discretization strategy inspired by the finite element method(DGM), assigning local sub-networks as surrogates within each subdomain. As these local parameters are decoupled from one another, different cells can only communicate through weak constraint, flux, with their neighbors. Therefore, the training is more stable and avoids the global instabilities observed in single-network architectures.

\begin{remark}
     For the low-frequency solution, PINNs eventually achieve higher accuracy than all weak-form methods. This outcome arises because strong-form approaches constrain the second derivatives of the solution, while weak-form approaches relax the regularity of solution by integration by parts. However, this improved accuracy demands a greater computational cost, because high order terms needs backward, which is computational costly. In particular, when the solution features strong high-frequency perturbations, PINNs nearly fail to converge.
\end{remark}
\subsection{2D Poisson Equation}
\begin{figure}[htbp]
    \centering
    \begin{minipage}[t]{0.45\textwidth}  
    \centering
    \includegraphics[width=\textwidth]{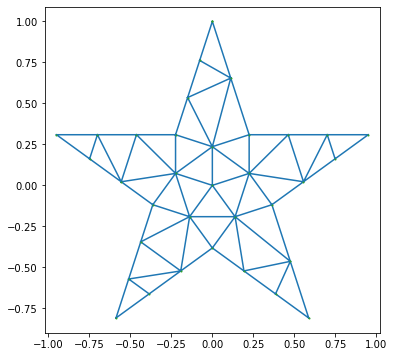}  
    \caption*{(a) The decomposition of $\Omega$}
    \label{decom}
  \end{minipage}
  \begin{minipage}[t]{0.45\textwidth} 
    \centering
    \includegraphics[width=\textwidth]{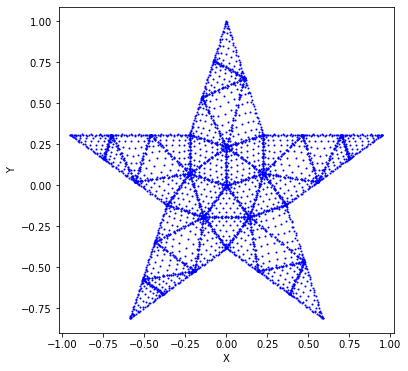} 
    \caption*{(b) The integration points}
    \label{polymesh}
  \end{minipage}
    \caption{(a) An example of decomposition of pentangle into the non-uniform mesh. In this case, the domain $\Omega$ is divided into 38 elements, each with an area of at least 0.05. The overall structure consists of 35 grid points and 71 edges. (b) The overall estimation points. In each element $E$, we take 15 integration points, while on each edge of $\partial E$, we take 20 integration points.}
    \label{polygon}
\end{figure}

Then, we consider the 2D Poisson Equation on an irregular domain $\Omega$. The equation is as follows:
\begin{equation}\label{2dPoisson}
\left\{
    \begin{aligned}
        &-\Delta u =f \qquad x\in \Omega\\
        &u = 0\qquad x\in \partial \Omega
    \end{aligned}\right.
\end{equation}
where $f=10$ and the domain is chosen as a pentangle with homogeneous Dirichlet boundary condition. The difficulty of this equation is dealing with the irregular domain. To solve this problem, we triangulate the domain just like DGM or FEM. The domain $\Omega$ and an example of its decomposition are shown in Figure \ref{polygon}(a). Then, we define test functions on each element as in Algorithm \ref{alg2}. The overall integration points are shown in Figure \ref{polygon}(b). 

In our setup, we choose the DNN-based solution for each element as MLP with $l=2, N_{hidden}=50$. To evaluate the weak form residuals, we take $N_{E}=15$ inner integration points and $N_e = 20$ points on each edge as Figure \ref{polygon}(b), summing up to 1990 points.
We set $deg=3$ for test polynomials. For comparison, we set MLP($l=3, N_{hidden}=128$) for vanilla PINN, ResNet($l=4, N_{hidden}=128$) for DeepRitz. To estimate the loss for PINN and DeepRitz, we choose 13726 points within the polygon domain.

\begin{figure}[ht]
    \centering
        \begin{minipage}[t]{0.45\textwidth}  
    \centering
    \includegraphics[width=\textwidth]{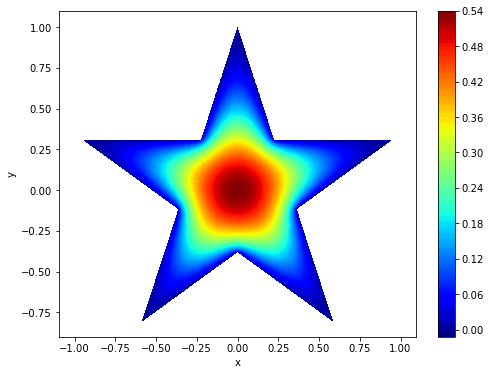}  
    \caption*{(a) Reference solution}
  \end{minipage}
  \begin{minipage}[t]{0.45\textwidth} 
    \centering
    \includegraphics[width=\textwidth]{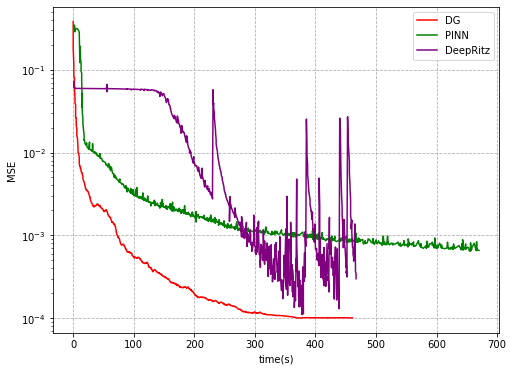} 
    \caption*{(b) MSE vs time(s)}
  \end{minipage}
  
  \begin{minipage}[t]{\textwidth}  
    \centering
    \includegraphics[width=\textwidth]{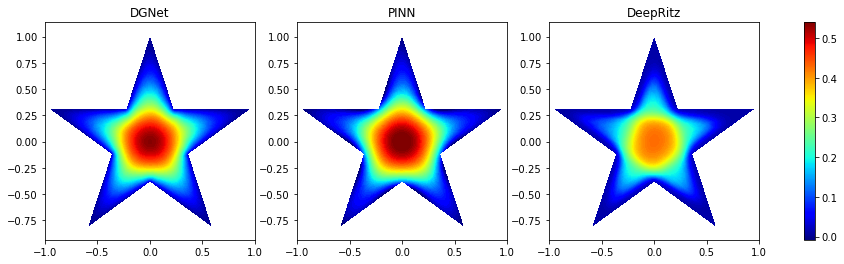}  
    \caption*{(c) The prediction solution}
  \end{minipage}

  \begin{minipage}[t]{\textwidth}  
    \centering
    \includegraphics[width=\textwidth]{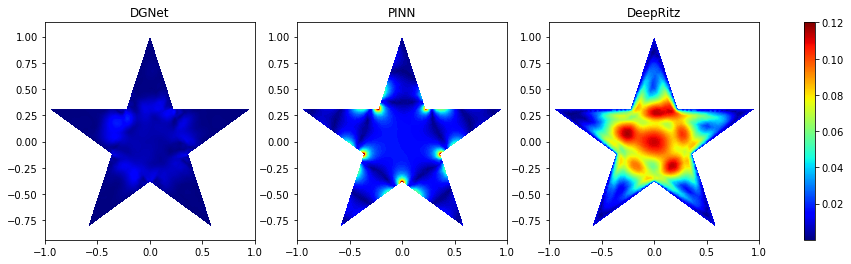}  
    \caption*{(d) Point-wise error}
  \end{minipage}
    \caption{The performance of different methods in solving Poisson equation on irregular doamain. (a) The reference solution. (b) Mean Square Error(MSE) vs. computation times. (c) The solution $u_{NN}$ solved by different methods. (d) The point-wise error $\left|u_{NN} - u_{exact}\right|$ obtained by different methods.}
    \label{2dPoissonexperiments}
\end{figure}

The experimental results are depicted in Figure \ref{2dPoissonexperiments}. 
Figure \ref{2dPoissonexperiments}(b) demonstrates that DGNN is capable of rapidly reducing the error to \(10^{-4}\) while maintaining a highly stable training process. In contrast, although PINN exhibits a decreasing error trend, its convergence is significantly slower. Compared to DGNN, the DeepRitz method, which optimizes a global energy functional, is highly susceptible to local loss perturbations affecting global parameters. As a result, its training process exhibits substantial fluctuations, leading to instability.  

Figure \ref{2dPoissonexperiments}(d) further illustrates that DGNN effectively addresses singularity-induced issues. While PINN performs well over a large domain, it inevitably suffers from substantial errors near singularities. The DeepRitz method, on the other hand, can achieve a high level of accuracy by leveraging prolonged training and selecting the optimal model. Nevertheless, its accuracy remains inferior to that of DGNN.  

\begin{remark}
    In this case, hpVPINN was not included in the comparative experiments. According to the original hpVPINN paper, the method addresses irregular boundaries not by employing an unstructured network but through an adaptive rectangular mesh partitioning strategy to handle singularities. This significantly increases computational complexity, which is why hpVPINN was excluded from the comparison.
\end{remark}

\subsection{Burgers Equation}
We proceed to solve the 1D Burgers Equation with periodic boundary conditions:
\begin{equation}\label{1dburgers}
    \left\{
    \begin{aligned}
        &\frac{\partial u}{\partial t}+u\frac{\partial u}{\partial x} = 0, \qquad x\in \Omega=(0,2\pi), t\in [0, 1.5]\\
        &u(x, 0)=u_0(x),\qquad x\in \Omega=(0,2\pi)\\
        &u(0, t) = u(2\pi, t); u_x(0, t) = u_x(2\pi, t),\qquad t\in [0, 1.5]
    \end{aligned}
    \right.
\end{equation}
where $u_0(x)=sin(x)+0.5$. Equations of this type are particularly challenging to solve due to the presence of discontinuities in their solutions. A representative example is Burgers' equation, which exhibits similar discontinuous behavior. Typically, Burgers' equation is studied with an added diffusion term that smooths extreme solutions, in \citep{raissi2017physicsinformeddeeplearning}. However, in this work, we consider only the convection term, as the presence of diffusion would otherwise smooth out sharp features and diminish the severity of the discontinuities. The reference solution, which is computed as in Appendix..., is presented in Figure \ref{1dburgersexperiments} (a). In classical numerical methods such as FDM and FEM, discontinuous solutions will lead to numerical instabilities characterized by oscillations near discontinuities. In DGM, this instability is significantly mitigated, and post-processing techniques, such as incorporating limiters, are employed to effectively suppress spurious oscillations around discontinuities.

To solve this equation, DGNN adopts an approach similar to the discontinuous Galerkin method (DGM), seeking a weak solution within a piece-wise DNN function space. In DGNN, we set time points $N_t=30$, order of test function $deg=3$, and the spatial domain is partitioned into $N=11$ intervals, with $N_{int}20$ Gauss–Legendre quadrature points utilized within each interval. For each spatial cell, a MLP with $l=2, N_{hidden}=50$, is employed as the local surrogate model. To evaluate the strong residuals of the vanilla PINN, we generate 100,000 collocation points within the space-time domain. For this comparison, the vanilla PINN employs a multilayer perceptron (MLP) architecture consisting of 4 layers, each containing 128 neurons. For the hp-VPINN method, following the suggested settings, the time points $N_t=100$, and the spatial domain is divided into 3 subdomains. Each spatial subdomain utilizes 40 Gaussian-Legendre quadrature points, with the polynomial degree of the test functions set as deg = 10. 

Figure \ref{1dburgersexperiments} presents the numerical results based on the 1d Burgers' equation. From Figure \ref{1dburgersexperiments}(b), we can observe that DGNN achieves very high accuracy in an extremely short time. Combining Figure \ref{1dburgersexperiments}(c) and Figure \ref{1dburgersexperiments}(d), it can be seen that DGNN maintains high accuracy globally while preserving local conservation effectively.
In contrast, methods such as PINN, hpVPINN, and PWNN struggle to converge due to the presence of discontinuous solutions. Notably, for PINN, which relies on the strong form, the derivative $u_x$ tends to infinity at discontinuities, making the strong form ineffective. As a result, significant errors occur in the global domain.
For PWNN, after incorporating the time dimension, requires sampling spatial points at each time step for training. This leads to compute test functions every training iteration, leading to extremely slow training speed.
Similarly, for hpVPINN, neither increasing the number of test regions nor raising the order of test functions can match the speed and accuracy achieved by DGNN.

\begin{figure}[ht]
    \centering
        \begin{minipage}[t]{0.45\textwidth}  
    \centering
    \includegraphics[width=\textwidth]{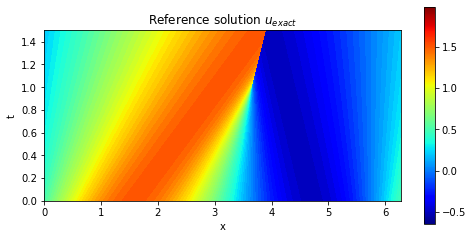}  
    \caption*{(a) Reference solution}
  \end{minipage}
  \begin{minipage}[t]{0.45\textwidth} 
    \centering
    \includegraphics[width=\textwidth]{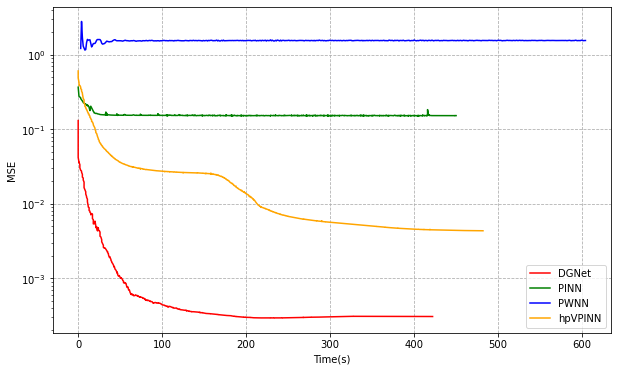} 
    \caption*{(b) MSE vs time(s)}
  \end{minipage}
  
  \begin{minipage}[t]{\textwidth}
    \centering
    \includegraphics[width=\textwidth]{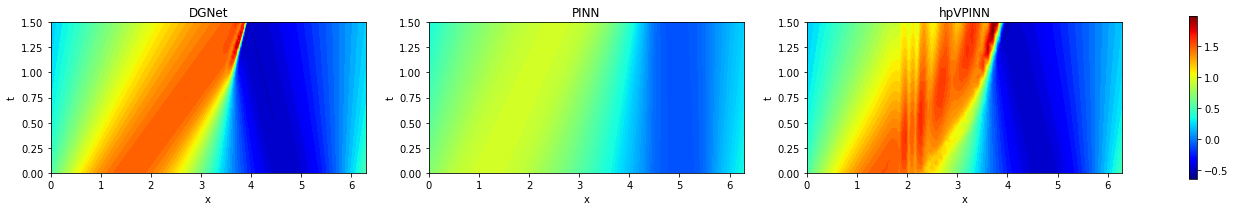}  
    \caption*{(c) The prediction solution}
  \end{minipage}

  \begin{minipage}[t]{\textwidth}  
    \centering
    \includegraphics[width=\textwidth]{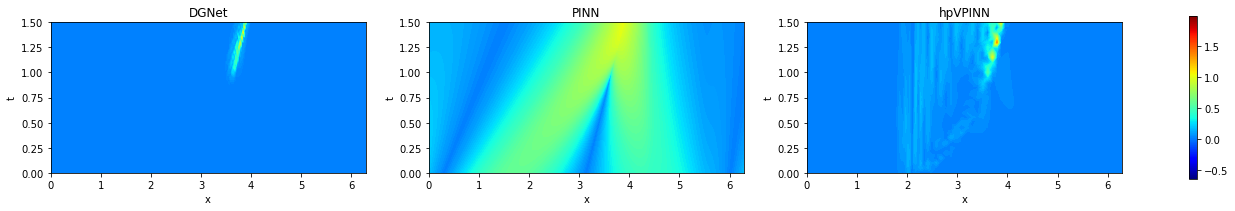}  
    \caption*{(d) Point-wise error}
  \end{minipage}
    \caption{The performance of different methods in solving 1D Burgers equation. (a) The reference solution. (b) Mean Square Error(MSE) vs. computation times. (c) The solutions $u_{NN}$ solved by different methods. (d) The point-wise error $\left|u_{NN} - u_{exact}\right|$ obtained by different methods.}
    \label{1dburgersexperiments}
\end{figure}
\begin{remark}
    It is worth noting that many existing methods, such as PWNN and PINN, can be seamlessly integrated into our framework. Specifically, different training strategies can be adopted locally, including varying test functions, test regions, and loss function designs. This flexibility corresponds to the concept of p-adaptivity in classical discontinuous Galerkin (DG) methods.
\end{remark}

\subsection{Additional ablations}
In this section, we delve into the impact of the varying hyperparameters on the DGNN, like $N_E, N_e, N, deg, l, N_{hidden}$. In Appendix \ref{integrate}, we have stated the conclusion that when $15\leq N_E\leq 20, N_e\geq 15, 0\leq deg \leq 10$, the exact solution has a loss at machine precision. Here we investigate the impact of mesh discretization on experimental results using the Poisson equation defined on irregular domains \ref{2dPoisson} as a case study. All parameters except the mesh partitioning settings follow the standardized configurations described earlier. We train each case for 20000 iterations by LBFGS optimizer. As shown in Table \ref{ablation}, the refinement level of meshes exhibits no significant impact on solution accuracy. Furthermore, our implementation of parallel linear layers ensures that increasing mesh density does not proportionally escalate training time. This design effectively circumvents the computational burden observed in PWNN, where the number of integration points—and consequently computational time—grows linearly with randomly sampled collocation points, thus preserving time efficiency while maintaining numerical robustness. 
\begin{table}[H]
    \centering
    \begin{tabular}{|c|c|c|c|c|c|}
    \hline
    \multicolumn{6}{|c|}{DGNN} \\
    \hline
     $s_{min}$ & 0.08 & 0.05 & 0.03 & 0.02 & 0.01 \\
     \hline
     N &    30 & 38 & 56 & 91 & 175 \\
    \hline
    MSE & 9.1383e-5 & 5.2068e-5 & 3.0097e-5 & 2.9410e-5 & 7.4054e-5\\
    \hline
    MAE & 5.87e-2 & 1.86e-2 & 2.42e-2 & 1.73e-2 & 1.94e-2 \\
    \hline
    Time(s)) &632.3 & 936.1 & 934.4 & 934.1 & 999.5 \\
    \hline
\end{tabular}
    \caption{Different triangulation of domain. Each module is MLP with $l=2, N_{hidden}=20$ and $deg = 3$.}
    \label{ablation}
\end{table}

\section{Conclusion and future works\label{conclusion}}
The novelty of this work lies in two interconnected contributions. First, we propose a neural network architecture that replaces traditional weight-sharing mechanisms with a \textbf{divide-and-conquer} strategy, where submodels are trained independently while weakly coupled through \textbf{numerical flux-based information transfer}. This design not only accelerates training convergence and enhances stability but also relaxes the regularity constraints on solutions, enabling efficient handling of discontinuous or oscillatory patterns. Second, we redefine the loss function by embedding \textbf{weak formulations} over local triangular elements, which inherently enforce physical conservation laws and stabilize element-wise solutions. The triangulation-driven discretization further reduces computational overhead by minimizing redundant evaluation points across complex domains. Experimental validations demonstrate that DGNN consistently outperforms state-of-the-art methods in resolving multi-scale phenomena, irregular geometries, and sharp discontinuities, achieving a robust balance between computational efficiency and solution fidelity while preserving physical interpretability.

\bibliographystyle{unsrtnat}  
\bibliography{references}  
\newpage
\appendix
\section{Exact solution of Burgers Equation}
Consider \ref{1dburgers}. Assume $x = x(t)$ s.t. 
\begin{equation}
    \begin{aligned}
        &x'(t) = u(x(t), t)\\
        &x(0) = x_0
    \end{aligned}
\end{equation}
then we have 
\begin{equation}
    \frac{d(u\left(x(t), t\right))}{dt}=\frac{\partial u}{\partial x}\frac{dx}{dt}+u_t=uu_x+ u_t=0
\end{equation}
Hence we have 
\begin{equation}
    \left\{\begin{aligned}
        & u(x(t, x_0), t) = C\\
        &u(x(0, x_0), 0)=g(x_0)
    \end{aligned}\right.
    \Rightarrow
    \left\{\begin{aligned}
        &u(x(t), t)=g(x_0)\\
        &x(0)=x_0
    \end{aligned}\right.
\end{equation}
which is the characteristic of  Burgers equation. Then $x(t)=g(x_0)t+x_0$. The Burgers Equation $u(x, t)$ can be rewritten as:
\begin{equation}
    \frac{du}{dx}=\frac{du}{dx_0}\frac{\partial x_0}{\partial x} = \frac{g'(x_0)}{\frac{\partial x}{\partial x_0}}=\frac{g'(x_0)}{1+g'(x_0)t}
\end{equation}
Therefore, when $1+g'(x_0)t=0$, $u_x\rightarrow \infty$. Hence, the time $t = -\frac{1}{g'(x_0)}\geq 0$ is the break time if exists. In Equation \ref{1dburgers}, $g(x)=sin(x)+\frac{1}{2}$. So the earliest breaking time $t_b=\min \left(\frac{1}{-cos(x_0)}\right)=1$ when $x_0=\pi$.

For the exact solution, we write the modified equation:
\begin{equation}\left\{
    \begin{aligned}
        &w = sin(x-wt-\frac{t}{2})\\
        &u=w+\frac{1}{2}
    \end{aligned}\right.
\end{equation}
and set the initial value $w = 0.8$ beyond the shock wave, i.e. $x < \pi + 0.5t$, the initial value $w = -0.8$ after the shock wave, i.e. $x > \pi + 0.5t$.

\section{Different choices of hyperparameters\label{integrate}}
\begin{figure}[H]
    \centering
    \begin{minipage}[t]{0.45\textwidth}  
    \centering
    \includegraphics[width=\textwidth]{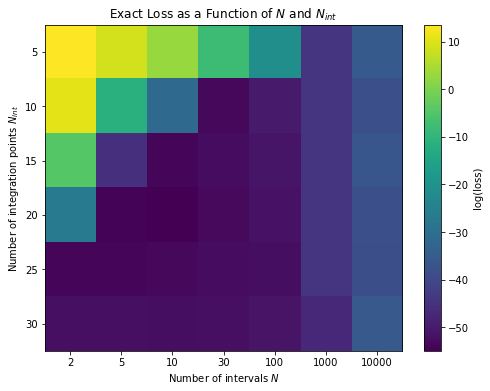}  
    \caption*{(a) Exact loss as a function of $N_x, N_{int}$}
  \end{minipage}
  \begin{minipage}[t]{0.45\textwidth} 
    \centering
    \includegraphics[width=\textwidth]{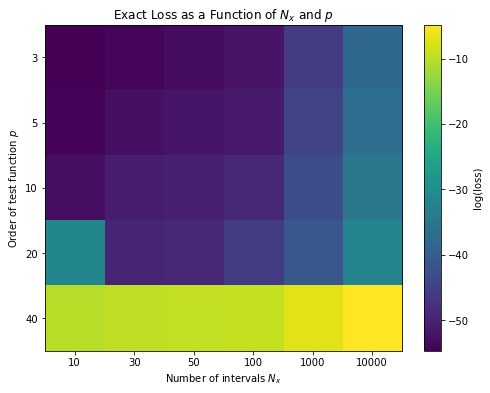} 
    \caption*{(b) Exact loss as a function of $N_x, p$}
  \end{minipage}
    \caption{Different choices of hyperparameters for case \ref{1dPoisson15pi}. (a) The exact loss as a function of $N_x$(the number of intervals) and $N_int$(the number of 1D Gaussian-Legendre quadrature points). (b) The exact loss as a function of $N_x$ and $p$(the order of test functions)}
    \label{diffchoices}
\end{figure}

First, we consider the 1d Poisson equation with $w=15\pi$. In Figure \ref{diffchoices}, We compute the loss of the exact solution as a function of $N_{int}$ and $N$. As we can see, for integration on edges, as long as $N_{e}\geq 15$, the error of the exact solution remains at machine precision. Therefore, $N_e$ is set to 15 or 20 in all the following experiments.

\begin{figure}[H]
    \centering
    \begin{minipage}[t]{0.45\textwidth}  
    \centering
    \includegraphics[width=\textwidth]{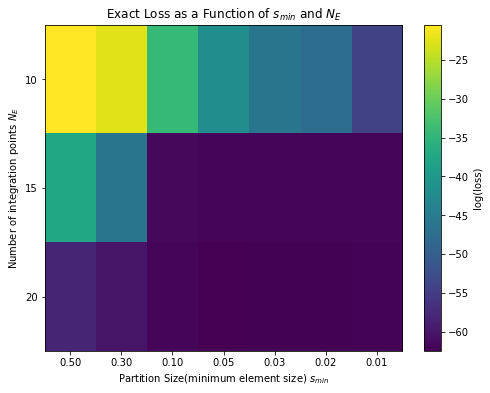}  
    \caption*{(a) Exact loss as a function of $N_E, s_{min}$}
  \end{minipage}
  \begin{minipage}[t]{0.45\textwidth} 
    \centering
    \includegraphics[width=\textwidth]{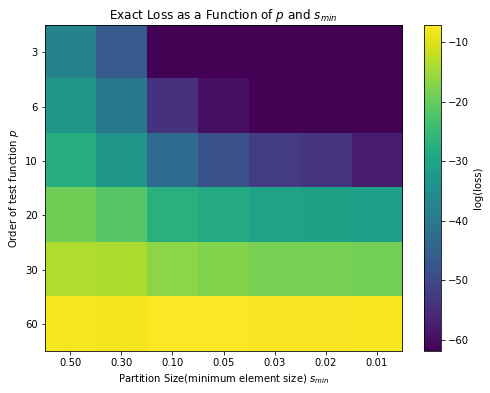} 
    \caption*{(b) Exact loss as a function of $N_E, p$}
  \end{minipage}
    \caption{Different choices of hyperparameters for case \ref{2dPoisson} in regular domain. (a) The exact loss as a function of $s_{min}$(the square of the minimum triangle element) and $N_int$(the number of 2D Gaussian-Legendre quadrature points). (b) The exact loss as a function of $s_{min}$ and $p$(the order of test functions)}
    \label{N_Eduibi}
\end{figure}

Then, we consider the integration number $N_E$ in elements. Take the 2D regular Poisson equation as an example. As is shown in Figure \ref{N_Eduibi}, $N_E$ set as 15 or 20 is enough for element integration. Hence, in the following experiments, we set $N_E=15$ for coarse mesh and $N_E=20$ for fine mesh. Meanwhile, the order of test functions $deg$ should not be greater than 10.





\end{document}